\documentclass[10pt,twocolumn,letterpaper]{article}
\usepackage[pagenumbers]{cvpr}

\usepackage{amsmath,amsfonts,bm}

\def\eqref#1{equation~\ref{#1}}

\def\1{\bm{1}}

\DeclareMathAlphabet{\mathsfit}{\encodingdefault}{\sfdefault}{m}{sl}
\SetMathAlphabet{\mathsfit}{bold}{\encodingdefault}{\sfdefault}{bx}{n}

\def\eqref#1{Eq.~(\ref{#1})}
\usepackage{graphicx}
\usepackage{float}
\usepackage{booktabs}
\usepackage{amsmath}
\usepackage{amsthm}
\newtheoremstyle{stmtit}{4pt}{4pt}{\itshape}{}{\bfseries}{.}{ }{\thmname{#1}\thmnote{ \normalfont(#3)}}
\newtheoremstyle{stmtup}{4pt}{4pt}{\normalfont}{}{\bfseries}{.}{ }{\thmname{#1}\thmnote{ \normalfont(#3)}}
\newcommand{\stmtname}{}
\theoremstyle{stmtit}\newtheorem*{stmtbody}{\stmtname}
\theoremstyle{stmtup}\newtheorem*{stmtdbody}{\stmtname}
\newenvironment{stmt}[2]{\renewcommand{\stmtname}{#1}\begin{stmtbody}[#2]}{\end{stmtbody}}
\newenvironment{stmtdef}[2]{\renewcommand{\stmtname}{#1}\begin{stmtdbody}[#2]}{\end{stmtdbody}}
\usepackage{multirow}
\usepackage{tabularx}
\usepackage{subcaption}
\usepackage{xcolor}
\usepackage{multicol}
\usepackage{wrapfig}
\usepackage{placeins}
\usepackage{algorithm}
\usepackage{algpseudocode}
\usepackage{microtype}
\definecolor{cvprblue}{rgb}{0.21,0.49,0.74}
\usepackage[pagebackref,breaklinks,colorlinks,allcolors=cvprblue]{hyperref}
\def\confName{CVPR}
\def\confYear{2027}
\title{CamCanvas: Absolute Camera Placement in Video World Models}
\author{Haoyang Chen \qquad Caiwen Ding\textsuperscript{\dag}\\University of Minnesota, Twin Cities}

\newcommand{\second}[1]{\underline{#1}}

\graphicspath{{figures/}{./}}
\providecommand{\linenumbers}{}
\begin{document}
\twocolumn[{\renewcommand\twocolumn[1][]{#1}%
\maketitle
\begin{center}
\vspace{-1.2em}
\includegraphics[width=\textwidth]{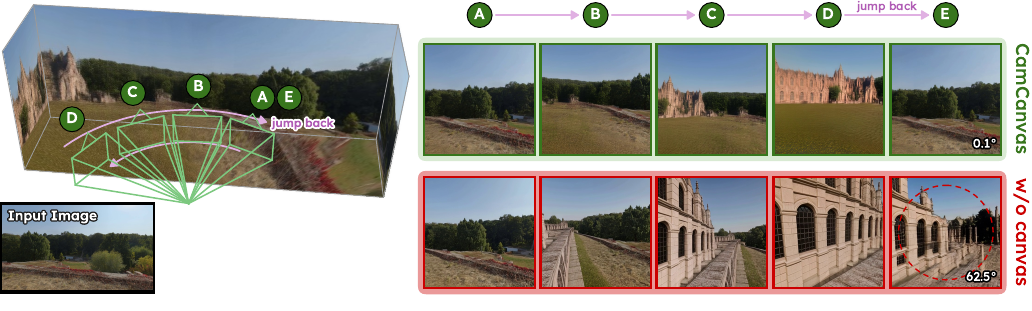}
\captionsetup{type=figure}
\caption{\textbf{Jumping back to a view generated long ago.} Left: content CamCanvas generated from the input image, with camera directions A to E. After sweeping from A to D, the camera jumps back to A (E). CamCanvas returns within $0.1^\circ$, while the backbone without the canvas lands $62.5^\circ$ away.}
\label{fig:teaser}
\end{center}
\vspace{0.6em}
}]
\begin{abstract}{\renewcommand{\thefootnote}{\fnsymbol{footnote}}\footnotetext[2]{Corresponding author.}}%
Interactive video world models generate a world chunk by chunk as camera commands arrive.
A command should place the camera at the requested pose at once, and a revisited place should look as it did.
Current models meet neither requirement: they follow small camera steps but execute only a fraction of a large jump, and a camera commanded back to a place visited long ago shows a different view.
We trace this failure to relative pose encodings and a finite attention window, and show that such a generator cannot recover a first visit that has left the window.
Our key insight is that returning to a visited pose is a matter of retrieving content rather than executing a motion.
We present CamCanvas, which stores the model's own generated world outside the attention window, as a panorama under rotation and a point cloud under translation, keeping the first observation of each location.
For every chunk, it renders this canvas at the commanded pose and supplies the view as a noise-free condition bound to each frame.
To evaluate this capability, we introduce the return test, which commands the camera away and back and measures, from the frames alone, the viewing-angle difference between the return and the model's own first visit.
CamCanvas raises the return AUC from $0$ to $94.7$ on $90^\circ$ jump returns, from $0.5$ to $96.2$ on real video, and from $0$ to $96.3$ in real indoor scenes, and transfers to SANA-WM, a world model with a different architecture.
\end{abstract}
\section{Introduction}\label{sec:intro}

Video world models are becoming interactive environments, generating the world chunk by chunk as camera commands arrive \citep{alonso2024diamond,valevski2025gamengen,worldplay,matrixgame,lingbotworld}.
Each camera command is a query: ``put me at this pose and show me what is there.''
People explore in this way, shifting gaze and head in rapid jumps while the world remains stable \citep{wurtz2008neuronal,freedman2008coordination}, and virtual reality offers snap turning and teleportation as standard controls \citep{kelly2025fov,prithul2021teleportation}.
An interactive world model should meet two requirements: a command places the camera at the requested pose at once, and a place the camera returns to looks as it did.

Current streaming world models meet neither requirement (Fig.~\ref{fig:teaser}).
They follow camera commands of a few degrees per step but execute only a small fraction of the rotation on jumps of $51$--$110^\circ$ (Sec.~\ref{sec:return:base}).
When commanded back to a place visited long ago, Wan2.2-5B ends a median $70^\circ$ from the view it generated there on $90^\circ$ returns. HY-WorldPlay, which retrieves stored views, shows a plausible view at the wrong pose and ends a median $81.4^\circ$ away.
The missing capability is \emph{absolute camera placement}: the commanded pose, not the path taken to it, determines which part of the generated world is on screen.

We trace this failure to how camera commands are encoded: commands are relative to the frames in view \citep{prope,viewrope,ucm}, and the model attends only to a finite window of recent frames, so once a place has left the window nothing in the input identifies what was generated there.
We prove that a generator conditioned only on window frames and relative commands cannot recover a first visit beyond what the window retains (Theorem~1, Sec.~\ref{sec:return:blindness}). Better camera encodings or longer training cannot close this gap.
Memory that retrieves earlier views \citep{worldmem,vmem,contextmemory,spmem} brings content back, but as views taken at other poses, leaving the model to work out the requested one.
Methods that render a 3D cache at the target view, such as GEN3C and Voyager \citep{gen3c,voyager}, supply the requested view directly, but generate offline along a trajectory given in advance.
Recent streaming variants render the cache as commands arrive, yet refresh it with new outputs \citep{zhao2026geostream} or inject the warped view as a partially noised condition \citep{kong2025worldwarp}.
Consequently, a return reads content that has drifted since the first visit, or a condition that the model regenerates rather than reproduces.

Our key insight is that returning to a visited pose is a matter of retrieving content rather than executing a motion.
Achieving this in interactive generation requires three properties.
First, the stored content must preserve the first visit.
Second, it must reach the model intact and at the commanded pose.
Third, each frame must read only the view rendered for its own pose.
We present CamCanvas, which satisfies all three (Fig.~\ref{fig:method}).
Specifically, CamCanvas stores the model's own generated world outside the attention window, as a panorama under rotation and a point cloud under translation, and writes each location only once.
For every chunk, it renders this canvas at the commanded pose and supplies the result as a noise-free condition aligned with the frames to be generated, binding each frame to its own rendered view.
Our experiments confirm that each property is necessary: refreshing the stored content, noising the condition, or removing the binding lowers the return AUC from $93.5$ to $4.7$, $9.7$, and $68.2$, respectively.

To measure this capability, we introduce the return test, which commands the camera away and back and compares the return with the model's own first visit, from the frames alone and without aligning trajectories (Sec.~\ref{sec:return-test}).
On $90^\circ$ jump returns, CamCanvas raises the return AUC, the area under the cumulative error curve up to $5^\circ$, from $0$ to $94.7$ and reduces the median return error from $70^\circ$ to $0.2^\circ$, close to the pose estimator's floor. GEN3C, which also renders a 3D cache, reaches $30.0$.
The AUC rises from $0.5$ to $96.2$ on real-video jump returns and from $0$ to $96.3$ in real indoor scenes, and the same condition remains effective on SANA-WM \citep{sanawm}, a model with a different architecture.
As a result, a camera command can place the view anywhere in a world the model has generated, not only next to where the camera currently is.

In summary, our contributions are as follows:
\begin{itemize}
\item We identify absolute camera placement as a requirement of interactive world models and prove that relative commands with a finite window cannot meet it once a place has left the window (Theorem~1).
\item We propose CamCanvas, which achieves placement by retrieving content rather than executing motion, and identify three properties this requires: first-visit storage, an intact condition at the commanded pose, and per-frame binding.
\item We introduce the return test, which scores a jump return against the model's own first visit from the frames alone and exposes failures that trajectory-averaged camera metrics hide.
\end{itemize}

\section{Related Work}\label{sec:related}

\textbf{Interactive video world models and camera control.}
World models predict an environment's response to actions~\citep{ha2018worldmodels,bruce2024genie}, and block-causal video models do so chunk by chunk, responding to camera commands chosen during exploration~\citep{chen2024diffusionforcing,causvid,selfforcing,selfforcingpp,rollingforcing,longlive,matrixgame,worldplay,lingbotworld,solarwm}.
Camera conditioning represents motion through Pl\"ucker rays, relative projective encodings, and discrete actions~\citep{motionctrl,cameractrl,ac3d,cameractrlii,prope,matrixgame,viewrope,ucm}.
These representations support camera control, but relative encodings specify movement from visible context without locating the camera among previously generated scene content~\citep{prope,solarwm}.
CamCanvas retains this camera conditioning and adds a rendered view that ties the commanded pose to content in the generated world.

\textbf{Long-horizon consistency and memory.}
Interactive world models report preserving the appearance of places revisited during exploration~\citep{matrixgame3,lingbotworld}.
Memory modules store past views and retrieve them by time or spatial location~\citep{worldmem,contextmemory,vmem,spmem,huang2025memoryforcing,mosaicmem,wang2026anchorweave,worldplay}.
These methods extend access to earlier content beyond the attention window, but a retrieved view was observed at another pose, so a return frame can preserve scene appearance without placing the camera at the commanded pose.
CamCanvas instead renders stored content at the commanded pose.

\textbf{Rendered conditions and geometric caches.}
GEN3C and related methods render stored geometry at target views, providing direct visual evidence that improves camera control and geometric consistency~\citep{gen3c,viewcrafter,trajectorycrafter,uni3c,voyager,hu2026geometry}.
GEN3C and Voyager also extend their caches with generated frames~\citep{gen3c,voyager}.
These methods generate offline along trajectories specified in advance, so outbound frames are generated knowing that the camera will come back.
Interactive generation instead receives commands one at a time, without knowing the future trajectory, and a generated chunk is final.
Recent streaming methods bring rendered geometry online.
GeoStream rebuilds its cache from its own outputs at each refresh~\citep{zhao2026geostream}, while WorldWarp warps cached content into new views as a partially noised condition~\citep{kong2025worldwarp}.
OctWorld fuses generated frames and depth into an octree~\citep{octworld}, and MoVerse renders a scaffold lifted from a panorama completed before exploration~\citep{moverse}.
A panorama completed in advance holds views from a single camera center and is fixed before any command arrives, whereas interactive commands arrive during generation and may translate the camera.
Both of our canvases keep each location's first observation instead of refreshing or fusing it, so a return reads what the first visit generated.
CamCanvas supplies the rendered view as a noise-free condition aligned with the target pose (Sec.~\ref{sec:camcanvas:injection}), and Table~\ref{tab:design-compare} compares these design choices.

\textbf{Evaluating camera control and returns.}
Camera benchmarks align generated and reference trajectories to remove scale or reference-frame differences~\citep{camerabench,worldscore,cami2v}, and long-horizon benchmarks assess content preservation and drift~\citep{worldroambench,wbench}.
Revisit protocols compare return frames with earlier observations through image similarity~\citep{vmem,worldplay}, which mixes viewpoint error with content change~\citep{worldroambench} and rewards a camera that barely moves~\citep{r2mbench}.
Cycle and forth-and-back trajectories bring the camera back along the path it took~\citep{vmem,hu2026geometry}, so the return is reached through small relative steps.
Our return test instead commands jumps back to poses whose content has left the attention window. It uses the pose estimator VGGT~\citep{vggt} to estimate the viewing-angle difference between the return frame and the model's own first-visit frame, without aligning trajectories or comparing with a reference video.

\section{CamCanvas}
\label{sec:camcanvas}
CamCanvas supplies the view at the commanded pose, so that the model need not infer where the camera is from relative motion.
This raises three questions: what to store so that a return can show its first visit (Sec.~\ref{sec:camcanvas:state}), how to supply it so that the model reproduces it rather than infers it, and which stored view each frame reads (Sec.~\ref{sec:camcanvas:injection}).
Figure~\ref{fig:method} gives an overview, and Algorithm~\ref{alg:inference} in Appendix~\ref{app:implementation} summarizes generation.

\begin{figure*}[t]
\centering
\includegraphics[width=\textwidth]{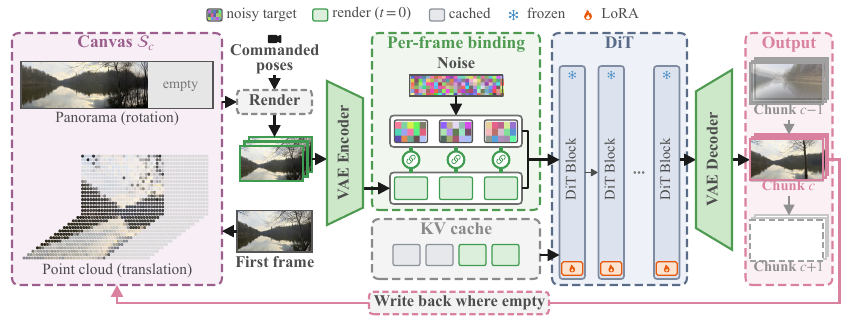}
\caption{\textbf{Overview of CamCanvas.} For every chunk, the canvas is rendered at each commanded pose and placed after the noisy targets at timestep zero, with target $k$ attending only to render $k$. The decoded chunk is written back only where the canvas is still empty.}
\label{fig:method}
\end{figure*}

\subsection{Preliminaries}
\label{sec:preliminaries}
We build on block-causal Wan2.2-5B \citep{wan2025,solarwm}, which denoises three latent frames $z_c$ per chunk in four steps and decodes them into pixel frames $x_c$.
Its video variational autoencoder (VAE) maps $1+4n$ pixel frames to $1+n$ latent frames and encodes the first frame $y_0$ separately.
The key--value (KV) window $\mathcal W_c$ holds up to 18 latent frames without an attention sink~\citep{xiao2024streamingllm}, so the first frame is evicted like any other frame.
Camera commands enter through PRoPE \citep{prope,solarwm}: the tokens of latent frame $k$ carry $P_k=\operatorname{lift}(K)W_k$, where $W_k\in\mathrm{SE}(3)$ is the commanded world-to-camera pose and $K$ holds fixed normalized intrinsics lifted to homogeneous coordinates.
As a result, attention depends on poses only through relative products:
\begin{equation}
 \ell_{ij}=\mathbf q_i^\top(P_iP_j^{-1})\mathbf k_j,\qquad
 \mathbf o_i=\sum_j a_{ij}(P_iP_j^{-1})\mathbf v_j,
 \label{eq:prope}
\end{equation}
where $i,j$ index frames, suppressing token indices; $\mathbf q_i,\mathbf k_j,\mathbf v_j$ are query, key, value vectors, $\ell_{ij}$ the logit, $a_{ij}$ its softmax weight, and $\mathbf o_i$ the output.
For chunk indices $\mathcal C_c$, let $u_c=\{P_iP_j^{-1}:i\in\mathcal C_c,\ j\in\mathcal W_c\}$. Each chunk is then sampled from
\begin{equation}
 p_\theta(z_c\mid z_{\mathcal W_c},u_c),
 \label{eq:chunk-conditional}
\end{equation}
where $\theta$ denotes the model parameters and $z_{\mathcal W_c}$ the clean context latents in the KV window.

\subsection{Returns and Return Blindness}
\label{sec:return:definitions}
\begin{stmtdef}{Definition 1}{absolute camera placement}Absolute camera placement is the ability to put the camera at a commanded pose in the generated world, so that the pose alone determines which previously generated content is visible.\end{stmtdef}
\begin{stmtdef}{Definition 2}{return}A chunk returns when its commanded pose is within $10^\circ$ and $0.15\bar D$ of a pose visited at least two chunks earlier, after the camera has departed by at least $25^\circ$ or $0.4\bar D$. Here $\bar D$ is the largest distance from the first commanded camera center to any center issued so far, so the rule depends only on past commands.\end{stmtdef}
We call an instantaneous change of the commanded pose between chunks a \emph{camera jump}.
Let $y_j$ and $y_c$ denote the model's first-visit and return frames, with $y_c$ the frame of the last latent in the returning chunk.
With a relative-rotation estimator $E(y,y')\in\mathrm{SO}(3)$ and the geodesic angle $\angle$ from the identity, the \emph{return error} $e=\angle E(y_j,y_c)$ is the orientation difference between the return and the model's own first visit, estimated from the frames without alignment.

\label{sec:return:blindness}
\begin{stmt}{Theorem 1}{return blindness}For any generator of the form \eqref{eq:chunk-conditional} and a first visit $j\notin\mathcal W_c$, the return satisfies $x_c\perp y_j\mid(z_{\mathcal W_c},u_c)$.\end{stmt}
The proof is in Appendix~\ref{app:theory:blindness}. Once the window no longer holds the first visit, its content must be supplied from outside the window.

\subsection{Writing and Reading the Canvas}
\label{sec:camcanvas:state}
A return must show what the first visit generated, so the canvas keeps each location's first observation and never overwrites it. It is kept outside the attention window: adding every stored frame to attention would grow its cost with the number of frames, while a fixed window would still bound access to earlier content (Theorem~1).
A single panorama cannot hold views taken from different camera centers, so the canvas takes one of two forms.
Under pure rotation every view is part of one sphere of directions, and the canvas is an equirectangular panorama. Under translation, it is a point cloud.
Both are initialized from the first frame and extended with the decoded outputs of every chunk.

\textbf{Rotation canvas.} The state before chunk $c$ is $\mathcal S_c=(C_c,M_c)$, with panoramic content $C_c$ and a binary observed mask $M_c$. We reproject each decoded chunk $x_c$ onto the panorama to obtain content $\widetilde x_c$ and an in-view indicator $B_c$, and update pixelwise:
\begin{equation}
\begin{aligned}
 U_c&=B_c(1-M_c),\qquad M_{c+1}=\max(M_c,U_c),\\
 C_{c+1}&=(1-U_c)C_c+U_c\widetilde x_c.
\end{aligned}
 \label{eq:write}
\end{equation}
The update mask $U_c$ admits only previously unobserved locations, so each return reads content whose error is incurred once, when it is written and read.

\textbf{Translation canvas.} The state is a point set $\mathcal P_c$ of colored 3D points. For every new chunk we estimate depth with VGGT \citep{vggt} and align its scale to the stored points. As in the panorama, pixels where the stored points already render content are discarded, and the remaining pixels are unprojected at the commanded pose and added to $\mathcal P_c$:
\begin{equation}
 \mathcal P_{c+1}=\mathcal P_c\cup\big\{\pi^{-1}_{W_k}(u,\,s_c\,d_k(u)),\ x_k(u)\big\}_{u\in V_k},
 \label{eq:write-depth}
\end{equation}
where $d_k$ is the estimated depth of frame $k$, $s_c$ the aligned scale, $V_k$ the retained pixels, and $\pi^{-1}_{W}$ unprojection at pose $W$.
Appendix~\ref{app:implementation} gives the alignment and rejection rules and the RealEstate10K setting.

\textbf{Reading.} For each pixel frame $p$ of a chunk, we render $r_p=\Pi(\mathcal S_c,W^{\mathrm{pix}}_p)$, where $\Pi$ is the renderer and $W^{\mathrm{pix}}_p$ is the commanded pose of that pixel frame, obtained from the sequence $W_k$. Rotations resample the panorama on the sphere of directions. Translations splat the stored points with a depth buffer, so nearer points occlude farther ones.
The rendered view covers a fraction $\kappa$ of the target, and the generator completes the remaining $1-\kappa$.
Consequently, newly written pixels come from a frame generated with the covered part of the same view as its condition, so content on both sides of a boundary between stored and new content is produced together in one frame.

\subsection{Injecting the Commanded View}
\label{sec:camcanvas:injection}
The rendered view and the frames to be generated share the commanded pose, so their relative transform in \eqref{eq:prope} is the identity.
This alignment lets attention reproduce the requested view directly, without inferring a change of viewpoint.

Rather than adding rendered views as extra context frames, we encode the rendered sequence with the target's video VAE and interleave condition and target tokens on the same token grid, which keeps command following intact (Sec.~\ref{sec:ablation}). Each chunk holds its three noisy target latents followed by their three condition latents, and each condition token carries its target's pose.
The condition is supplied for every chunk, not only for returns, since a jump can come in any chunk and the model receives no separate signal for when a return occurs.
The condition tokens stay at timestep zero, since noise would remove information needed for reproduction.

\textbf{Per-frame binding.} Within a chunk, target latent $k$ attends to its own condition latent and not to the conditions of the other latents in the chunk. History, the other target latents, and all condition tokens of earlier chunks remain visible.
Without this binding, latent $k$ also sees the views rendered for the other poses of its chunk, so the content it reproduces is not tied to a single view.

\begin{table*}[t]
\centering
\caption{\textbf{Panorama jump returns.} AUC: area under the curve of the fraction of returns with $e$ below a threshold, for thresholds up to $5^\circ$ (\%); $e$: median return error ($^\circ$); $\phi$: out-of-window fraction. Best in bold, second best underlined.}
\label{tab:jumpback}
\small\setlength{\tabcolsep}{3.2pt}
\begin{tabular}{l*{10}{c}}
\toprule
 & \multicolumn{2}{c}{within window} & \multicolumn{2}{c}{$51^\circ$} & \multicolumn{2}{c}{$70^\circ$} & \multicolumn{2}{c}{$90^\circ$} & \multicolumn{2}{c}{$110^\circ$} \\
 & \multicolumn{2}{c}{$\phi=2\%$} & \multicolumn{2}{c}{$\phi=23\%$} & \multicolumn{2}{c}{$\phi=45\%$} & \multicolumn{2}{c}{$\phi=60\%$} & \multicolumn{2}{c}{$\phi=77\%$} \\
\cmidrule(lr){2-3}\cmidrule(lr){4-5}\cmidrule(lr){6-7}\cmidrule(lr){8-9}\cmidrule(lr){10-11}
Method & AUC$\uparrow$ & $e\downarrow$ & AUC$\uparrow$ & $e\downarrow$ & AUC$\uparrow$ & $e\downarrow$ & AUC$\uparrow$ & $e\downarrow$ & AUC$\uparrow$ & $e\downarrow$ \\
\midrule
\textit{Ground-truth renders} & \textit{93.8} & \textit{0.2} & \textit{95.0} & \textit{0.2} & \textit{96.1} & \textit{0.1} & \textit{96.8} & \textit{0.1} & \textit{97.7} & \textit{0.1} \\
\midrule
Wan2.2-5B & 0.0 & 53.8 & 1.6 & 60.4 & 0.0 & 70.0 & 0.0 & 70.0 & 0.0 & 53.5 \\
MiniMax-H3-33B & 5.0 & 19.0 & 0.0 & 24.1 & 6.8 & 18.9 & 28.8 & \underline{3.5} & 48.4 & \underline{2.0} \\
LingBot-World & 0.2 & 64.6 & 0.0 & 44.3 & 0.0 & 62.8 & 0.0 & 74.2 & 1.7 & 58.1 \\
HY-WorldPlay & 0.0 & 39.3 & 0.0 & 42.7 & 0.0 & 63.5 & 0.0 & 81.4 & 0.0 & 124.1 \\
GEN3C & \underline{68.0} & \underline{0.6} & \underline{69.4} & \underline{0.4} & \underline{50.3} & \underline{1.2} & \underline{30.0} & 36.8 & \underline{48.5} & 2.6 \\
CamCanvas & \textbf{89.3} & \textbf{0.3} & \textbf{87.3} & \textbf{0.2} & \textbf{94.1} & \textbf{0.3} & \textbf{94.7} & \textbf{0.2} & \textbf{96.0} & \textbf{0.2} \\
\bottomrule
\end{tabular}
\end{table*}

\begin{table*}[t]
\centering
\caption{\textbf{Returns on RealEstate10K.} PSNR and LPIPS compare returns with the real frames. Best in bold, second best underlined.}
\label{tab:realvideo}
\small\setlength{\tabcolsep}{3.5pt}
\begin{tabular}{lcccccccc}
\toprule
 & \multicolumn{4}{c}{Jump return to the first frame} & \multicolumn{4}{c}{Out and back} \\
\cmidrule(lr){2-5}\cmidrule(lr){6-9}
Method & AUC$\uparrow$ & $e\downarrow$ & PSNR$\uparrow$ & LPIPS$\downarrow$ & $\bar e\downarrow$ & PSNR$\uparrow$ & LPIPS$\downarrow$ & MEt3R$\downarrow$ \\
\midrule
Wan2.2-5B & 0.5 & 25.2 & 9.9 & 0.703 & 6.3 & 10.7 & 0.650 & 0.329 \\
HY-WorldPlay & 18.0 & 8.7 & 10.1 & 0.707 & 13.0 & 9.7 & 0.725 & 0.291 \\
LingBot-World & 5.2 & 13.3 & 12.6 & 0.576 & 27.4 & 10.8 & 0.715 & 0.414 \\
GEN3C & \textbf{98.4} & \textbf{0.0} & \textbf{32.1} & \textbf{0.057} & \textbf{0.1} & \textbf{16.5} & \textbf{0.379} & \textbf{0.037} \\
CamCanvas & \second{96.2} & \second{0.2} & \second{21.0} & \second{0.199} & \second{0.5} & \second{14.9} & \second{0.430} & \second{0.140} \\
\bottomrule
\end{tabular}
\end{table*}

\subsection{Training}
\label{sec:camcanvas:training}
We keep the block-causal Wan2.2-5B world model frozen and adapt only its self-attention \citep{wan2025,solarwm}, so that the backbone's generation is kept and the adapter learns to read the condition. We attach rank-32 low-rank adapters~\citep{hu2022lora} to the query, key, value, and output projections of all 30 self-attention blocks (23.6M trainable parameters), while cross-attention and feed-forward layers stay frozen.

We train the adapter with teacher-forced flow matching \citep{lipman2023flow} on 21-latent-frame clips, with the per-frame binding as at inference. Each three-frame chunk samples its own diffusion timestep $t$ and is interleaved at timestep zero with its rendered condition $r_c$, which keeps only content observed in the first frame or in earlier chunks, rendered at the chunk's commanded poses, as the canvas holds at inference. With $z_c^t=(1-t)z_c+t\epsilon$ and $\epsilon\sim\mathcal N(0,I)$, the loss supervises the flow velocity on noisy target tokens only:
\begin{equation}
 \mathcal L(\psi)=\mathbb E_{c,t,\epsilon}\Big[w(t)\,\big\|v_{\theta,\psi}(z_c^t,t\mid z_{<c},u_c,r_c)-v^\star_c\big\|_2^2\Big],
 \label{eq:objective}
\end{equation}
where $\psi$ denotes the adapter parameters, $w(t)$ the scheduler weight, $z_{<c}$ the ground-truth context, and $v^\star_c$ the target velocity (Appendix~\ref{app:implementation}). Training runs for 2000 steps. Rotation clips come from about 900 CC0 Poly Haven panoramas~\citep{polyhaven} at $2048\times4096$ and include camera jumps and jumps back to earlier views. Translation clips come from TartanAir and Sekai, with conditions built by per-chunk depth write-back as at inference \citep{tartanair,sekai}.

\section{Experiments}
\label{sec:experiments}
We first describe the setup (Sec.~\ref{sec:setup}). We then examine whether existing world models execute jumps and return to their first visit, and whether CamCanvas returns across rotation, translation, real video, and indoor scenes (Sec.~\ref{sec:main-results}). Finally, we ablate its design choices (Sec.~\ref{sec:ablation}).

\subsection{Setup}
\label{sec:setup}
\textbf{Protocols.}\label{sec:return-test}
We evaluate whether models reach the commanded viewpoint and preserve content by commanding the camera away and back, using the protocols detailed in Appendix~\ref{app:protocols}.
The panorama protocol tests four jump sizes ($51^\circ$--$110^\circ$) and a within-window return on 91 panoramas, with 249 frames per trajectory (Fig.~\ref{fig:teaser}).
Two protocols use 300 clips of the RealEstate10K test set \citep{zhou2018stereo}: a jump return to an earlier real frame and an out-and-back trajectory.
Rotation and translation cycles, in which the camera moves out and retraces its path back, use the same 91 panoramas.
Partial-coverage returns command a return whose target view the canvas covers only in part, so that the generator must complete the rest.
Indoor returns use 40 rendered trajectories in ten real HM3D buildings \citep{hm3d}, where the camera looks away and back with ground-truth renders available for every frame.
We use jumps because they separate reaching the commanded viewpoint from smooth tracking and occur in practice as gaze shifts, snap turns, cuts, and navigation goals \citep{land2004,farmani2018,cutting2010,savva2019habitat}.

\textbf{Metrics.}
The return test measures the viewing-angle difference $e$ between the return and the model's own first visit, estimated by VGGT without alignment \citep{vggt}. We report its median and the return AUC, the area under the curve of the fraction of returns with $e$ below a threshold, for thresholds from $0$ to $5^\circ$, in percent. Ground-truth renders reach $94$--$98$, the estimator's floor.
The out-of-window fraction $\phi$ is the share of the target view outside the attention window.
PSNR and LPIPS~\citep{zhang2018lpips} compare returns with the real frames on RealEstate10K, which tests agreement with the observed scene, and with the paired first-visit frames on cycles.
We also report MEt3R \citep{met3r} for content consistency and mean return error $\bar e$ on cycles and out-and-back trajectories.
Under translation, orientation alone does not show whether the camera returned to the correct position, so we also report the position error $\bar d$. It is the distance between the estimated camera centers of each return frame and its paired outbound frame, normalized by the largest estimated distance from the first camera center and averaged over the return.
For partial-coverage returns we report the return AUC and fidelity on covered pixels against the first visit.

\textbf{Methods.}
We compare CamCanvas with the Wan2.2-5B block-causal base model \citep{wan2025,solarwm}, MiniMax-H3-33B \citep{solarwm}, LingBot-World \citep{lingbotworld}, HY-WorldPlay with retrieval-based memory \citep{worldplay}, and GEN3C, a bidirectional video model that renders a 3D cache into every frame \citep{gen3c}.
All models receive the same commands and first frames through their own inference code.
GEN3C runs its released code (Cosmos-7B) with our per-frame poses and intrinsics and MoGe depth on the first frame. It receives the whole trajectory before generation, which it splits into segments of 121 frames, so it knows each return command while generating the outbound frames.
Wan2.2-5B and CamCanvas share the backbone, prompts and seeds.
All results are averaged over three random seeds.

\subsection{Main Results}
\label{sec:main-results}
\begin{figure}[t]
\centering
\includegraphics[width=\columnwidth]{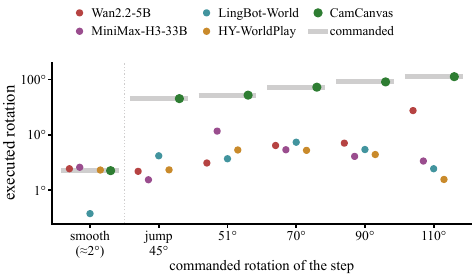}
\caption{\textbf{Existing models do not jump.} Median executed rotation per step; grey bars mark the commands. All models follow smooth steps, but only CamCanvas executes jumps.}
\label{fig:exec-rotation}
\end{figure}

\textbf{Existing models.}\label{sec:return:base}
Existing models follow small smooth commands but rotate far less than commanded on jumps (Fig.~\ref{fig:exec-rotation}).
Wan2.2-5B, MiniMax-H3-33B and HY-WorldPlay follow commands of about $2^\circ$ per step, while LingBot-World turns $0.4^\circ$.
Their median jump rotations are $1.6$--$11.7^\circ$, except for Wan2.2-5B's $27.4^\circ$ at $110^\circ$, while the median rotation of CamCanvas is within $1.3^\circ$ of the command at every level.
Across four backbones from three families, the return test exposes failures to reach the commanded viewpoint (Table~\ref{tab:jumpback}), and SANA-WM \citep{sanawm} has a median return error of $4.6^\circ$ even on the rotation cycle, where the camera returns smoothly.
Wan2.2-5B ends a median $54$--$70^\circ$ from its first visit at every level, and within-window returns fail as often as out-of-window returns.
Camera-following metrics averaged over trajectories made mostly of small steps can remain favorable while returns fail, even when the target remains in the window.

\begin{figure*}[t]
\centering
\includegraphics[width=\textwidth]{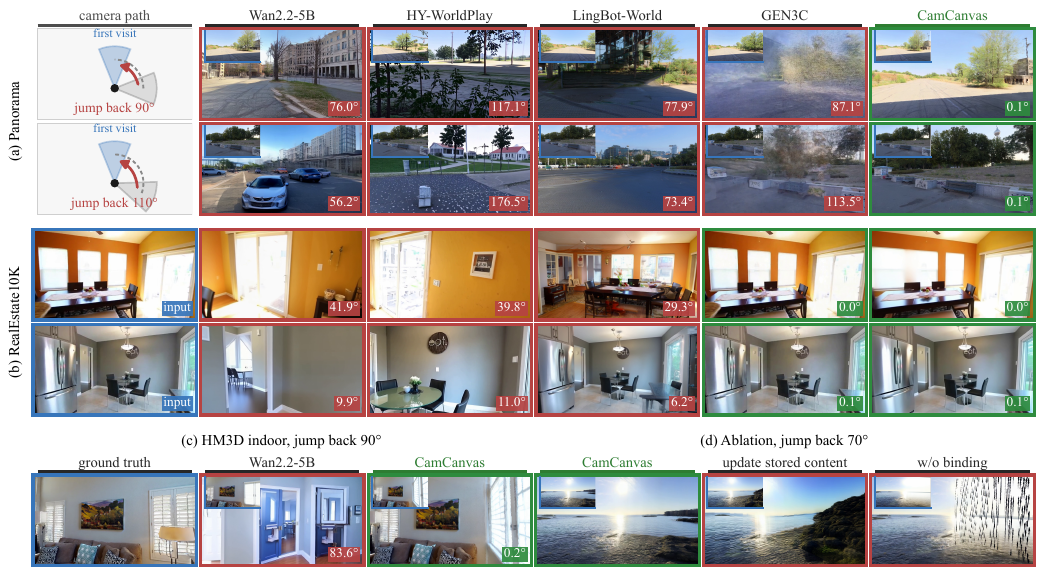}
\caption{\textbf{Qualitative returns.} (a) Panorama and (b) RealEstate10K jump returns, (c) an HM3D indoor return, and (d) ablation variants. Each return shows the method's own first visit as a blue inset (in (b), the input) and is framed green when its return error, in the corner, is below $5^\circ$ and red otherwise; in (d), red marks returns whose content differs from the first visit. Baselines land on a different view, while CamCanvas shows its first visit again. Updating stored content returns to a changed view, and removing per-frame binding tears the return.}
\label{fig:qualitative}
\end{figure*}

\begin{figure}[t]
\centering
\includegraphics[width=\columnwidth]{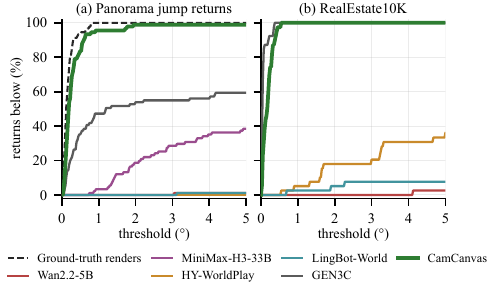}
\caption{\textbf{Return accuracy.} Fraction of returns whose return error is below a threshold; the area under each curve up to $5^\circ$ is the return AUC. (a) The 91 out-of-window panorama jump returns over all jump levels, with ground-truth renders (dashed) as the estimator floor. (b) RealEstate10K returns to the input frame.}
\label{fig:auc-curve}
\end{figure}

\textbf{Panorama jump returns.}
CamCanvas returns to its own first visit at every jump size, close to the estimator's floor: its return AUC is $87.3$--$96.0$ against $93.8$--$97.7$ for ground-truth renders and at most $1.7$ for Wan2.2-5B, LingBot-World and HY-WorldPlay (Table~\ref{tab:jumpback}, Fig.~\ref{fig:auc-curve}).
Those models end $39$--$124^\circ$ away from their own first visit, and HY-WorldPlay produces a plausible view at the wrong pose (Fig.~\ref{fig:qualitative}).
MiniMax-H3-33B keeps its first frame in attention, so its AUC rises only as the target moves closer to that frame, not because the camera reaches the commanded pose.
GEN3C reaches $30.0$--$69.4$, with $30.0$ at the $90^\circ$ level.
Within-window returns, whose target remains in attention, fail as well: the streaming baselines reach at most $5.0$ and GEN3C $68.0$, while CamCanvas reaches $89.3$.

\textbf{Real video.}
The same holds on real video, where a return must also reproduce a real frame. CamCanvas reaches a return AUC of $96.2$ on jump returns to the first frame against $0.5$ for the base model (Table~\ref{tab:realvideo}), with PSNR $21.0$ against $9.9$\,dB and LPIPS $0.199$ against $0.703$.
The streaming and retrieval-based baselines reach $5.2$--$18.0$ on the same jump return.
GEN3C reaches $98.4$ and reproduces the first frame more closely, while CamCanvas reaches $96.2$ generating chunk by chunk.
On out-and-back trajectories, CamCanvas reduces $\bar e$ to $0.5^\circ$ from the base model's $6.3^\circ$, compared with $13.0$--$27.4^\circ$ for those baselines, and GEN3C reaches $0.1^\circ$ with return PSNR $16.5$\,dB. CamCanvas reaches return PSNR $14.9$ against $10.7$\,dB for the base model and MEt3R $0.140$ against $0.329$.
Appendix~\ref{app:results} reports returns to real frame 3 (Table~\ref{tab:realvideo-f3}) and content consistency along return paths.

\begin{table}[t]
\centering
\caption{\textbf{Rotation and translation cycles.} Metrics compare paired return and outbound frames; $\bar d$ is the relative position error ($0.06$ on ground-truth renders). Best in bold, second best underlined.}
\label{tab:cycles}
\footnotesize\setlength{\tabcolsep}{1.6pt}
\resizebox{\columnwidth}{!}{%
\begin{tabular}{lccccccccc}
\toprule
 & \multicolumn{4}{c}{Rotation cycle} & \multicolumn{5}{c}{Translation cycle} \\
\cmidrule(lr){2-5}\cmidrule(lr){6-10}
Method & PSNR$\uparrow$ & LPIPS$\downarrow$ & MEt3R$\downarrow$ & $\bar e\downarrow$ & PSNR$\uparrow$ & LPIPS$\downarrow$ & MEt3R$\downarrow$ & $\bar e\downarrow$ & $\bar d\downarrow$ \\
\midrule
Wan2.2-5B & 15.1 & 0.408 & 0.225 & 6.6 & 15.8 & 0.463 & 0.320 & 7.9 & 0.384 \\
HY-WorldPlay & 11.2 & 0.568 & 0.265 & 27.5 & 14.4 & 0.467 & \second{0.296} & 4.2 & 0.305 \\
LingBot-World & 23.4 & 0.164 & 0.070 & 22.1 & 17.2 & \second{0.394} & 0.312 & 13.9 & \second{0.225} \\
GEN3C & \second{25.2} & \second{0.130} & \second{0.060} & \textbf{0.5} & \second{17.8} & 0.430 & 0.310 & \second{2.0} & 0.230 \\
CamCanvas & \textbf{27.9} & \textbf{0.092} & \textbf{0.044} & \second{0.6} & \textbf{24.3} & \textbf{0.156} & \textbf{0.083} & \textbf{0.4} & \textbf{0.036} \\
\bottomrule
\end{tabular}}
\end{table}

\begin{figure}[t]
\centering
\includegraphics[width=\columnwidth]{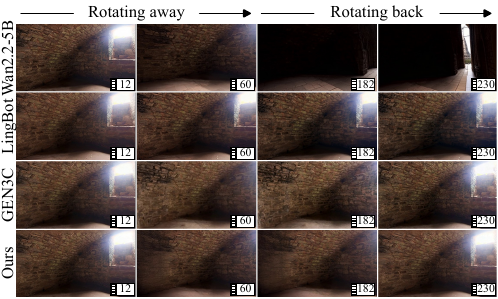}
\caption{\textbf{Rotation cycle.} Frames 182 and 230 return to the poses of frames 60 and 12. Wan2.2-5B returns to a different place, LingBot-World barely rotates, and GEN3C and CamCanvas return to the first-visit view.}
\label{fig:cycle-qual}
\end{figure}
\textbf{Rotation and translation cycles.}
Content scores alone reward a camera that barely moves, so we read them together with the return error.
On the rotation cycle (Table~\ref{tab:cycles}, Fig.~\ref{fig:cycle-qual}), CamCanvas reaches paired-frame PSNR $27.9$\,dB against $15.1$\,dB for Wan2.2-5B, MEt3R $0.044$ against $0.225$, and a mean return error of $0.6^\circ$ against $6.6^\circ$.
HY-WorldPlay retrieves stored views, yet its $\bar e$ is $27.5^\circ$.
LingBot-World reaches high content scores only because its camera turns $1.7^\circ$ of the commanded $40^\circ$.
On the translation cycle, CamCanvas lowers the mean return error from $7.9^\circ$ to $0.4^\circ$ and the position error $\bar d$ from $0.38$ to $0.036$, with paired-frame PSNR $24.3$\,dB against $15.8$\,dB.
GEN3C reaches a mean return error of $2.0^\circ$ and $\bar d$ of $0.23$, on par with LingBot-World's $0.22$, whose $\bar e$ is $13.9^\circ$. Its paired-frame PSNR is $17.8$\,dB. Fig.~\ref{fig:return-curve} plots PSNR along the return.

\textbf{Command following and completion.}
The canvas helps before any return as well: the median outbound orientation error is $2.08^\circ$ against $7.84^\circ$ for Wan2.2-5B (Table~\ref{tab:ablation}).
Where the canvas covers only part of the return target ($69\%$ on the median clip), the generator completes the rest: CamCanvas reaches a return AUC of $72.1$ with a median error of $0.76^\circ$ and reproduces the covered pixels at PSNR $26.3$\,dB against the first visit.

\textbf{Indoor scenes.}
On the HM3D look-back trajectories, the return AUC is $96.3$ for CamCanvas against $0$ for Wan2.2-5B, with median return error $0.13^\circ$ against $53.6^\circ$.
On 26 HM3D trajectories that orbit an object by $120^\circ$ and then jump back, CamCanvas ends at a median position return ratio, the remaining distance to the first camera center relative to the largest distance along the orbit, of $0.013$ against $0.472$ for Wan2.2-5B, with median rotation error $0.8^\circ$ against $36.8^\circ$ (Table~\ref{tab:orbit}).

\textbf{Transfer to another architecture.}
\begin{table}[t]
\centering
\caption{\textbf{SANA-WM transfer on the rotation cycle.} Metrics as in Table~\ref{tab:cycles}. Best results are in bold.}
\label{tab:sana}
\small\setlength{\tabcolsep}{3.5pt}
\begin{tabular}{lcccc}
\toprule
 & \multicolumn{4}{c}{Rotation cycle, 91 panoramas} \\
\cmidrule(lr){2-5}
Method & PSNR$\uparrow$ & LPIPS$\downarrow$ & MEt3R$\downarrow$ & $\bar e\downarrow$ \\
\midrule
SANA-WM & 15.4 & 0.425 & 0.253 & 8.0 \\
SANA-WM + CamCanvas & \textbf{23.2} & \textbf{0.196} & \textbf{0.118} & \textbf{1.4} \\
\bottomrule
\end{tabular}
\end{table}
CamCanvas also improves the rotation cycle on SANA-WM \citep{sanawm}, raising PSNR from $15.4$ to $23.2$\,dB, lowering LPIPS from $0.425$ to $0.196$ and MEt3R from $0.253$ to $0.118$, and reducing $\bar e$ from $8.0^\circ$ to $1.4^\circ$ (Table~\ref{tab:sana}).
SANA-WM uses linear attention~\citep{katharopoulos2020linear} and $8\times$ temporal compression. The same rendered condition carries over, supplied with a rank-32 adapter on the returning chunks that the rule of Sec.~\ref{sec:return:definitions} selects from the commands (Appendix~\ref{app:implementation}).

\subsection{Ablations}
\label{sec:ablation}
Table~\ref{tab:ablation} replaces one design choice at a time on a fixed subset of 30 panorama trajectories with the same returning chunks (Table~\ref{tab:appendix-return-schedules}). The choices concern what is stored, how it is supplied, and which view each frame reads.

\textit{What is stored.} A return can only show what the canvas still holds. Updating stored content with every observation, as caches refreshed from the model's outputs do~\citep{zhao2026geostream}, makes each return read content that has passed through more generations, and the return AUC drops from $93.5$ to $4.7$. It also degrades the outbound path: each chunk overwrites the region it sees with its own output, placed at the commanded pose, so a small deviation of one chunk is passed to the next, and the outbound error grows steadily to $22.46^\circ$. Keeping the canvas for only one chunk drops the AUC to $16.9$. Storing only the first frame and the last chunk, the cache kept by GEN3C's released code, reaches $90.7$, since these targets lie within the first $66^\circ$ of the sweep. On 30 returns to a view generated $114^\circ$ into a faster sweep, neither the first frame nor recent, its AUC falls to $2.6$ against $76.0$ for the full canvas.

\textit{How it is supplied.} The condition must carry the stored content intact and at the target pose. Noising it to the target's timestep, as in partially noised warped conditions~\citep{kong2025worldwarp}, drops the AUC to $9.7$. Supplying the rendered views as extra context frames lowers it to $54.9$ and loosens command following, with the outbound error growing from $2.08^\circ$ to $10.42^\circ$.

\textit{Which view each frame reads.} Without the per-frame binding, the AUC drops to $68.2$ and LPIPS against the first visit rises from $0.088$ to $0.262$: a latent that also sees the views rendered for other poses mixes them, and the return tears (Fig.~\ref{fig:qualitative}d).

\begin{table}[t]
\centering
\caption{\textbf{Ablations on panorama jump returns.} AUC and $e$ as in Table~\ref{tab:jumpback}; LPIPS compares returns with first visits; $e_{\text{out}}$: median outbound error ($^\circ$). Best in bold, second best underlined.}
\label{tab:ablation}
\small\setlength{\tabcolsep}{3.2pt}
\resizebox{\linewidth}{!}{%
\begin{tabular}{lcccc}
\toprule
Method & AUC$\uparrow$ & $e\downarrow$ & LPIPS$\downarrow$ & $e_{\text{out}}\downarrow$ \\
\midrule
Wan2.2-5B & 0.0 & 63.92 & 0.673 & 7.84 \\
CamCanvas & \textbf{93.5} & \textbf{0.22} & \textbf{0.088} & \textbf{2.08} \\
\midrule
Update instead of first write & 4.7 & 8.16 & 0.347 & 22.46 \\
Condition noised to target timestep & 9.7 & 11.14 & 0.505 & 3.85 \\
Canvas kept for one chunk & 16.9 & 21.84 & 0.463 & 2.76 \\
Canvas as context frames & 54.9 & 1.67 & 0.289 & 10.42 \\
First frame and last chunk only & \second{90.7} & \second{0.32} & \second{0.169} & 2.58 \\
Without per-frame binding & 68.2 & 1.29 & 0.262 & \second{2.42} \\
\bottomrule
\end{tabular}}
\end{table}

\section{Conclusion}\label{sec:conclusion}

We presented CamCanvas, which obtains absolute camera placement by retrieving the content generated at the commanded pose instead of executing the motion to it.
Relative commands with a finite window cannot recover a first visit that has left the window (Theorem~1), so CamCanvas keeps each location's first observation outside the window, supplies it intact at the commanded pose, and binds each frame to its own rendered view.
Measured by the return test against the model's own first visit, returns reach the commanded pose in panoramas, real videos, and indoor scenes, and on SANA-WM, a world model with a different architecture.
More broadly, these results suggest that an interactive world model can return to any place it has generated by retrieving its own outputs, rather than by attending to a longer history.

\textbf{Limitations and future work.} CamCanvas assumes static scenes: the canvas stores content as first generated, so moving objects would be shown where they were. Trajectories reach 249 frames, and translations stay at room scale. The translation canvas inherits monocular depth errors, which alignment and rejection limit but do not remove. Longer explorations, larger translations, and dynamic content remain future work.

{
\small
\bibliographystyle{ieeenat_fullname}
\bibliography{iclr2027_conference}
}

\clearpage
\onecolumn
\begin{center}
{\Large\textbf{Appendix}}
\end{center}
\vspace{1em}
\appendix
\setcounter{table}{0}\setcounter{figure}{0}\setcounter{algorithm}{0}\setcounter{equation}{5}
\renewcommand{\thetable}{S\arabic{table}}\renewcommand{\thefigure}{S\arabic{figure}}\renewcommand{\thealgorithm}{S\arabic{algorithm}}
\raggedbottom
The appendix is organized as follows. Sec.~\ref{app:protocols} specifies the evaluation protocols and how returns are scored. Sec.~\ref{app:implementation} gives training and implementation details. Sec.~\ref{app:results} reports additional quantitative and qualitative results, including GEN3C under interactive control and failure cases. Sec.~\ref{app:theory} states the assumptions and proofs of the theoretical results and measures the terms of Theorem~2.

\section{Evaluation Protocols and Scoring}
\label{app:protocols}

A command trajectory is $\mathcal T=(W_0,\dots,W_{L-1})$, where $W_k$ is the commanded world-to-camera pose at latent frame $k$ and $L$ is the number of latent frames. The attention window holds $N_w=18$ latent frames. A return at latent $m$ to target latent $j<m$ sets $W_m=\dots=W_{m+h-1}=W_j$, a camera jump kept for $h$ latent frames. For each return, the out-of-window fraction $\phi$ is the fraction of pixels in the target view whose scene direction was last in view at latent $\leq m-N_w$. It is computed from the commands alone before generation. The target age is $m-j$ latent frames.

\paragraph{Return rule.}
Definition~2 identifies returning chunks from the camera commands alone. We use the poses $W_k$ and chunk indices of Sec.~\ref{sec:preliminaries}, with indices starting at zero. Let $\mathbf c_k$ be the camera center obtained from $W_k^{-1}$. For chunk $c$, let $\bar D=\max_{k\leq 3c+2}\|\mathbf c_k-\mathbf c_0\|_2$ over the commands issued up to that chunk.

A chunk $c$ is a return when a commanded pose in that chunk is near an earlier pose $k$, and an intervening command departed from $k$. For three-latent-frame chunks, we test some $q\in\{3c,3c+1,3c+2\}$ and $k\leq 3c-6$. Near means rotational distance at most $10^\circ$ and camera-center distance at most $0.15\bar D$. Departure means that some $j$ with $k<j<3c$ has rotational distance at least $25^\circ$ or camera-center distance at least $0.4\bar D$ from $k$. When $\bar D\leq10^{-3}$, as on rotation-only trajectories, the translation terms are dropped. The departure condition distinguishes a return from continued observation during a slow sweep.

Table~\ref{tab:appendix-return-schedules} lists the returning chunks of each protocol. Because the rule reads only the commands, it is causal with respect to the generated frames and gives the same chunks for every method and ablation variant.

\begin{table}[H]
\centering
\begin{minipage}[t]{0.43\textwidth}
\centering
\caption{\textbf{Return schedules.} Chunks are zero-based. All schedules follow from Definition~2 and are shared by all variants. $^\dagger$Including the within-window return.}
\label{tab:appendix-return-schedules}
\small
\begin{tabular}{lc}
\toprule
Protocol & Returning chunks \\
\midrule
Panorama jump return$^\dagger$ & $18,19,20$ \\
RealEstate10K jump return & $7$ \\
Panorama rotation cycle & $15$--$20$ \\
Panorama translation cycle & $14$--$20$ \\
\bottomrule
\end{tabular}
\end{minipage}\hfill
\begin{minipage}[t]{0.53\textwidth}
\centering
\caption{\textbf{Panorama jump levels.} Target and age are given in latent frames. Each clip has one out-of-window return at latent 54 and one within-window return at latent 60; the within-window row covers the same 91 clips.}
\label{tab:appendix-jump-template}
\small
\setlength{\tabcolsep}{4pt}
\begin{tabular}{lrrrrr}
\toprule
Level & Target & Age & Jump ($^\circ$) & $\phi$ & Clips \\
\midrule
$51^\circ$ & 30 & 24 & 50.8 & 0.23 & 24 \\
$70^\circ$ & 21 & 33 & 70.4 & 0.45 & 23 \\
$90^\circ$ & 12 & 42 & 90.4 & 0.60 & 22 \\
$110^\circ$ & 3 & 51 & 110.1 & 0.77 & 22 \\
Within window, $45^\circ$ & 36 & 24 & 44.5 & 0.02 & 91 \\
\bottomrule
\end{tabular}
\end{minipage}
\end{table}

\paragraph{Panorama jump protocol.}
The outbound sweep changes yaw by $2.2^\circ$ per latent frame with a $4^\circ$ sinusoidal pitch oscillation. Each trajectory has $L=63$ latent frames, corresponding to 249 pixel frames, with one out-of-window return at $m=54$ kept for $h=3$ latents. Targets $j\in\{30,21,12,3\}$ are assigned cyclically across clips with random turn sign. These targets give the four out-of-window jump levels in Table~\ref{tab:appendix-jump-template}. Each clip also has a within-window return at $m=60$ to $j=36$, kept for $h=3$ latents. The protocol uses 91 CC0 Poly Haven panoramas not used in training \citep{polyhaven}. Each return is scored against that method's own first-visit frame at the target latent.

\paragraph{RealEstate10K jump protocol.}
Each trajectory has $L=24$ latents, corresponding to 93 pixel frames. Latents 0--20 follow the real trajectory. A camera jump at $m=21$, kept for $h=3$ latents, returns to the pose of real frame 0 in half of the clips and of real frame 3 in the other half, visited 21 and 17 latents earlier, respectively. The protocol uses 300 clips of the RealEstate10K test set \citep{zhou2018stereo}. We take CamI2V's 1,000 RealEstate10K test clips \citep{cami2v} in order and retain clips whose 16-frame camera path ends with displacement between 1.0 and 4.7 in RealEstate10K units and rotation between $5^\circ$ and $40^\circ$. We keep the first 300 qualifying clips, whose identifiers will be released with the code. Table~\ref{tab:realvideo} reports the 150 returns to real frame 0, and Table~\ref{tab:realvideo-f3} the 150 returns to real frame 3, whose first-visit frame is latent 4. The return frame $y_c$ is at latent 23. PSNR and LPIPS compare the 12 pixel frames of the returning chunk with the real frame at the target pose.

\paragraph{RealEstate10K out-and-back protocol.}
Each trajectory has $L=42$ latents. Latents 0--20 follow the real trajectory, and latents 21--41 revisit the same poses in reverse, with $W_k=W_{41-k}$ for $21\leq k\leq41$. This protocol uses the same 300 clips. Each return frame is paired with the outbound frame at the same commanded pose. Six return frames coincide with the poses of real frames 0, 3, 6, 9, 12 and 15, and PSNR and LPIPS compare them with those real frames. The translation canvas holds the first frame with MoGe depth \citep{moge}.

\paragraph{Panorama cycles.}
Each trajectory has $L=63$ latent frames, corresponding to 249 pixel frames. Outbound latents 0--31 are followed by return latents 32--62 along the outbound path in reverse. The same 91 panoramas each supply one rotation cycle and one translation cycle. Each return frame is paired with the outbound frame at the same commanded pose, following the pairing used by HY-WorldPlay and VMem \citep{worldplay,vmem}. Rotation cycles use the rotation canvas, and translation cycles use the translation canvas.

\paragraph{Partial-coverage protocol.}
The 91 panoramas and out-of-window returns of the panorama jump protocol are reused, but each return is commanded to the target orientation offset in pitch by $\pm15^\circ$, $\pm25^\circ$ or $\pm35^\circ$, so that the canvas covers only part of the return view, a median of $69\%$. Because the commanded return pose differs from the target, the return error is the angle between the executed rotation from the target latent to the last held latent, estimated by VGGT, and the commanded rotation. Fidelity is the PSNR on covered pixels between the return frames and the canvas rebuilt from the model's own frames before the return and rendered at the return poses.

\paragraph{HM3D look-back protocol.}
In each of ten HM3D buildings \citep{hm3d}, four camera positions are sampled $1.3$\,m above the floor, at least $0.6$\,m from any visible surface and with a median scene depth of at least $1.5$\,m, each with a random initial heading. Each of the 40 trajectories follows the panorama jump template (Table~\ref{tab:appendix-jump-template}), with jump level and turn direction cycled, and its out-of-window return is scored as in the panorama jump protocol. Ground-truth renders are available for every frame.

\paragraph{HM3D orbit protocol.}
In HM3D buildings \citep{hm3d}, a camera $1.3$\,m above the floor faces the surface point at the image center, $1.2$ to $3.0$\,m away. Over latents 0--32 it orbits this point on a horizontal circle by $120^\circ$ while facing it, and latents 33--41 jump back to the first pose and hold. Every orbit pose sees the point unoccluded. Returns are latents 34--41, compared with frame 0. The protocol uses 26 trajectories. Ground-truth renders are available for every frame. The position return ratio is the distance of the last estimated camera center from the first, divided by the largest distance from the first along the trajectory. The orbit error is the median over latents of the horizontal angle at the orbited point between estimated and commanded camera centers, after a least-squares scale fit per clip.

\paragraph{Pose estimation.}
VGGT-1B estimates the camera at each latent frame from the first pixel frame of that latent, with frame zero as the reference \citep{vggt}. Ground-truth renders pass through the same pipeline and give the estimator floor. VGGT also supplies the depth written into the translation canvas (Sec.~\ref{sec:camcanvas:state}). The rotation canvas and the translation canvas on RealEstate10K use no VGGT depth.

\FloatBarrier
\section{Training and Implementation}
\label{app:implementation}
Algorithm~\ref{alg:inference} summarizes generation with CamCanvas. This section then describes the backbone and adapter, the training objective and data, the canvas settings, and the adaptation to SANA-WM.

\begin{algorithm}[H]
\caption{Generation with CamCanvas}
\label{alg:inference}
\small
\begin{algorithmic}[1]
\Require First frame $y_0$, commands $W_k$ issued one chunk at a time, frozen generator $\theta$ with adapter $\psi$
\State Initialize the canvas $\mathcal S_0$ from $y_0$ (panorama, or point cloud with its depth)
\For{$c=0,1,\dots$}
 \State $r_p\gets\Pi(\mathcal S_c,W^{\mathrm{pix}}_p)$ for each pixel frame $p$ of chunk $c$ \Comment{render at the commanded pose}
 \State $z_c\sim p_{\theta,\psi}(z_c\mid z_{\mathcal W_c},u_c,r_c)$ \Comment{condition at timestep zero, per-frame binding}
 \State Decode $x_c$ and update $\mathcal S_{c+1}$ by \eqref{eq:write} or \eqref{eq:write-depth}
 \State Commit $[z_c\,|\,r_c]$ to the KV window
\EndFor
\end{algorithmic}
\end{algorithm}

\paragraph{Backbone and adapter.}
The backbone is a block-causal Wan2.2-5B distilled with Self Forcing from a bidirectional camera-conditioned model \citep{causvid,selfforcing,yin2024dmd2,solarwm}. It generates three latent frames per chunk with four denoising steps. Output images are $864\times480$ pixels, with horizontal and vertical fields of view of $89^\circ$ and $58^\circ$. For each adapted projection, the LoRA adapter~\citep{hu2022lora} adds $(\alpha/r)BA$ with $r=\alpha=32$. Matrix $A$ uses Kaiming initialization~\citep{he2015delving}, and $B$ starts at zero.

\paragraph{Conditioning and objective.}
Training uses teacher-forced flow matching on 21-latent-frame clips. A diffusion timestep is sampled for each three-frame chunk. Clean rendered conditions are interleaved as $[3\ \text{noisy}\mid3\ \text{condition}]$, producing 42 latent frames. Condition timesteps are zero, and each condition has the same pose as its corresponding target. The scheduler-weighted masked mean squared error supervises flow velocity on noisy latent frames only, with target
\begin{equation}
 v^\star_c=m_c\odot(\epsilon-z_c)+(1-m_c)\odot v_\theta(z_c^t,t\mid z_{<c},u_c),
 \label{eq:target}
\end{equation}
where $m_c$ marks the latent tokens whose $2\times2$ patch the render covers on average by at least half. Where the render has content, the target is the ground-truth velocity. Elsewhere it is the frozen backbone's prediction without the condition, since the condition does not determine that content. Condition frames and the supplied clean first frame are excluded from the loss. First-frame conditioning and the complete rendered condition each have dropout probability 0.1.

\begin{table}[H]
\centering
\begin{minipage}[t]{0.42\textwidth}
\centering
\caption{\textbf{Adapter optimization settings.}}
\label{tab:appendix-optimization}
\small
\begin{tabular}{ll}
\toprule
Training setting & Value \\
\midrule
Optimizer & AdamW~\citep{loshchilov2019adamw} \\
Learning rate & $10^{-4}$ \\
Betas & $(0.9,0.95)$ \\
Weight decay & 0.01 \\
Warm-up & 100 steps \\
Learning-rate schedule & Cosine with warm-up \\
Gradient norm clipping & 1 \\
Batch size & 1 clip \\
Precision & bfloat16 \\
Training duration & 2000 steps \\
\bottomrule
\end{tabular}
\end{minipage}\hfill
\begin{minipage}[t]{0.52\textwidth}
\centering
\caption{\textbf{Training mix.} Trajectory families in the mixed training set of 3366 clips, as sampled during training. Shares are rounded and sum to 101\%.}
\label{tab:appendix-training-mix}
\small
\begin{tabular}{lr}
\toprule
Trajectory family & Sampled share \\
\midrule
Translation with depth write-back & 32\% \\
Random walk & 15\% \\
Smooth sweep with jumps & 12\% \\
Jump & 12\% \\
Smooth sweep & 11\% \\
Jump back to an earlier view & 10\% \\
Large jump & 9\% \\
\bottomrule
\end{tabular}
\end{minipage}
\end{table}

\paragraph{Training data.}
Rotation trajectory families are smooth sweeps, random walks, jumps, and smooth sweeps with jumps. Training jump magnitudes are sampled uniformly from $10^\circ$ to $45^\circ$. For each target, the condition retains only content observed in the first frame or in earlier chunks, rendered at that target's command pose. Translation conditions are built by the inference write-back: the first frame enters with MoGe depth \citep{moge}, and later frames are written with VGGT depth aligned as in \eqref{eq:write-depth}, so the adapter sees the same kind of point-cloud render it reads at inference. Jump-back clips command a return to a view outside the window and hold it, and large jumps extend the jump range beyond that of the jump family. Table~\ref{tab:appendix-training-mix} gives the sampled proportions of the 3366 rotation and translation clips.

\paragraph{Canvas and inference.}
The rotation canvas is $8192\times16384$ pixels, in height-by-width order. It is initialized from the given first frame and updated from decoded generated frames, keeping the first observation of each location as in \eqref{eq:write}. The translation canvas is a point cloud initialized from the first frame with VGGT depth scaled to a median of three units, which fixes the unit of commanded translations, and extended by \eqref{eq:write-depth} only where the stored points render no content. Each new frame's depth is estimated jointly with the first frame and the last four written frames, so that it is consistent with what is already stored.

On RealEstate10K, the translation canvas holds the first frame with MoGe depth \citep{moge}, and pixels it leaves empty are filled from the rotation canvas. Alignment uses the median depth ratio on overlapping pixels after rejecting ratios beyond three scaled median absolute deviations. Frames with less than $5\%$ overlap or more than $10\%$ relative residual are not written. Pixels on depth discontinuities, where the depth range in a $3\times3$ window exceeds $10\%$ of its minimum, are dropped. Written masks are eroded by two pixels.

A read renders the canvas at the commanded pose of each pixel frame. For a rotation jump, the pixel frames of the jump latent are rendered at the post-jump pose rather than at interpolated intermediate poses, so the condition never shows a view the camera was not commanded to. Uncovered pixels are zero with an associated coverage mask.

\paragraph{Adaptation to SANA-WM.}
As for Wan2.2-5B, the condition is the canvas rendered at the commanded pose and encoded by the backbone's VAE \citep{sanawm}. SANA-WM uses linear attention and an LTX-2 VAE~\citep{hacohen2026ltx2} with eightfold temporal compression. Condition and target tokens are interleaved on the same token grid, with the per-frame binding and the objective of Sec.~\ref{sec:camcanvas:training}, and only rank-32 adapters are trained. The backbone is the chunk-causal SANA-WM checkpoint at $720$p with $1.6$B parameters. The adapters are trained for $2000$ steps at learning rate $10^{-4}$ on $16$-latent-frame panorama rotation clips with the backbone frozen.

\FloatBarrier
\section{Additional Results}
\label{app:results}
This section reports results referenced in Sec.~\ref{sec:experiments}. Table~\ref{tab:realvideo-f3} reports RealEstate10K jump returns to real frame 3, and Table~\ref{tab:orbit} the HM3D object orbits with a jump back (Sec.~\ref{sec:main-results}).

\begin{table}[H]
\centering
\begin{minipage}[t]{0.53\textwidth}
\centering
\caption{\textbf{Jump returns to real frame 3 on RealEstate10K.} Metrics as in Table~\ref{tab:realvideo}. Best in bold, second best underlined.}
\label{tab:realvideo-f3}
\small
\begin{tabular}{lcccc}
\toprule
Method & AUC$\uparrow$ & $e\downarrow$ & PSNR$\uparrow$ & LPIPS$\downarrow$ \\
\midrule
Wan2.2-5B & 1.6 & 18.4 & 10.3 & 0.690 \\
HY-WorldPlay & 13.8 & 8.7 & 10.0 & 0.695 \\
LingBot-World & 13.1 & 9.8 & 12.0 & 0.595 \\
GEN3C & \textbf{81.0} & \textbf{0.8} & \textbf{17.9} & \textbf{0.244} \\
CamCanvas & \second{76.4} & \second{1.0} & \second{17.3} & \second{0.285} \\
\bottomrule
\end{tabular}
\end{minipage}\hfill
\begin{minipage}[t]{0.44\textwidth}
\centering
\caption{\textbf{HM3D object orbits with a jump back.} $\rho_{\text{pos}}$: median position return ratio; $e_{\text{rot}}$: median rotation error of the return ($^\circ$); $e_{\text{orb}}$: median orbit error ($^\circ$); LPIPS compares returns with first visits.}
\label{tab:orbit}
\small\setlength{\tabcolsep}{3.5pt}
\begin{tabular}{lcccc}
\toprule
Method & $\rho_{\text{pos}}\downarrow$ & $e_{\text{rot}}\downarrow$ & $e_{\text{orb}}\downarrow$ & LPIPS$\downarrow$ \\
\midrule
Wan2.2-5B & 0.472 & 36.80 & 8.56 & 0.677 \\
CamCanvas & \textbf{0.013} & \textbf{0.80} & \textbf{1.89} & \textbf{0.366} \\
\bottomrule
\end{tabular}
\end{minipage}
\end{table}

\paragraph{GEN3C with interactive control.}
In Table~\ref{tab:realvideo}, GEN3C receives the whole trajectory in one 121-frame pass, so its outbound frames are generated with the return already known. We also run GEN3C interactively, following its official demo: the first pass receives only the outbound poses, and the second pass continues from the last generated frame with the return poses. Under interactive control, the AUC of GEN3C drops from $98.4$ to $92.7$ for returns to the first frame and from $81.0$ to $70.0$ for returns to real frame 3, and its PSNR for returns to the first frame drops from $32.1$ to $22.5$\,dB (Table~\ref{tab:gen3c-interactive}). CamCanvas, which generates as a stream, reaches AUC $96.2$ and $76.4$ on the two returns. Interactive GEN3C keeps a higher PSNR on returns to the first frame ($22.5$ against $21.0$\,dB), while CamCanvas is higher on returns to real frame 3 ($17.3$ against $15.6$\,dB).

\begin{table}[H]
\centering
\caption{\textbf{GEN3C with interactive control on 300 RealEstate10K jump clips.} Metrics as in Table~\ref{tab:realvideo}. Best in bold, second best underlined.}
\label{tab:gen3c-interactive}
\small\setlength{\tabcolsep}{3pt}
\begin{tabular}{lcccccccc}
\toprule
 & \multicolumn{4}{c}{Jump return to the first frame} & \multicolumn{4}{c}{Jump return to real frame 3} \\
\cmidrule(lr){2-5}\cmidrule(lr){6-9}
Method & AUC$\uparrow$ & $e\downarrow$ & PSNR$\uparrow$ & LPIPS$\downarrow$ & AUC$\uparrow$ & $e\downarrow$ & PSNR$\uparrow$ & LPIPS$\downarrow$ \\
\midrule
GEN3C, offline & \textbf{98.4} & \textbf{0.0} & \textbf{32.1} & \textbf{0.057} & \textbf{81.0} & \textbf{0.8} & \textbf{17.9} & \textbf{0.244} \\
GEN3C, interactive & 92.7 & \second{0.1} & \second{22.5} & \second{0.151} & 70.0 & 1.2 & 15.6 & 0.332 \\
CamCanvas & \second{96.2} & 0.2 & 21.0 & 0.199 & \second{76.4} & \second{1.0} & \second{17.3} & \second{0.285} \\
\bottomrule
\end{tabular}
\end{table}

\begin{figure}[H]
\centering
\includegraphics[width=\textwidth]{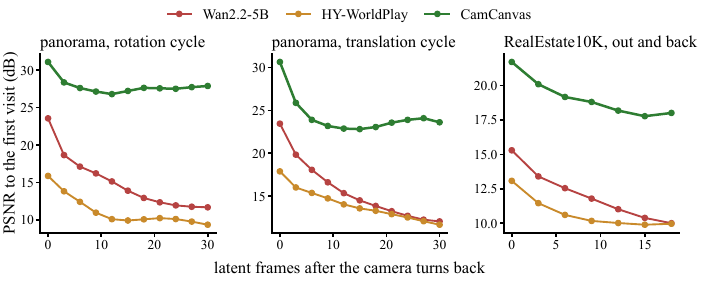}
\caption{\textbf{Content consistency along returns.} PSNR against the paired first-visit frames for Wan2.2-5B, HY-WorldPlay, and CamCanvas on the panorama rotation cycle, the panorama translation cycle, and RealEstate10K out-and-back trajectories. CamCanvas stays close to the first visit along the return, while the baselines keep drifting.}
\label{fig:return-curve}
\end{figure}

Table~\ref{tab:design-compare} compares the design choices of methods that condition on stored content (Sec.~\ref{sec:related}), and Fig.~\ref{fig:return-curve} plots PSNR against the first visit along the return paths of the cycle and out-and-back protocols.

\begin{table}[H]
\centering
\caption{\textbf{Design choices of methods that condition on stored content.} Interactive: commands arrive one at a time and generated frames are final. Stored content: what is kept and how it is updated. Condition: what the generator receives for the target view.}
\label{tab:design-compare}
\small\setlength{\tabcolsep}{3pt}
\begin{tabularx}{\linewidth}{@{}l c >{\raggedright\arraybackslash}X >{\raggedright\arraybackslash}X@{}}
\toprule
Method & Interactive & Stored content and update & Condition for the target view \\
\midrule
GEN3C \citep{gen3c} & no & point cloud; first frame plus the latest added frame & render, concatenated along channels \\
Voyager \citep{voyager} & no & point cloud; points added in unseen regions & RGB-D render \\
GeoStream \citep{zhao2026geostream} & yes & point cloud rebuilt from the latest frame at each refresh & render of the rebuilt cache \\
WorldWarp \citep{kong2025worldwarp} & yes & 3D Gaussians re-optimized per chunk & warped content, partially noised \\
HY-WorldPlay \citep{worldplay} & yes & past frames, retrieved & frames observed at other poses \\
CamCanvas & yes & rotation: panorama; translation: point cloud from per-chunk depth; both first write & render at the target pose, noise-free, per frame \\
\bottomrule
\end{tabularx}
\end{table}

\begin{figure}[!htb]
\centering
\includegraphics[width=\textwidth]{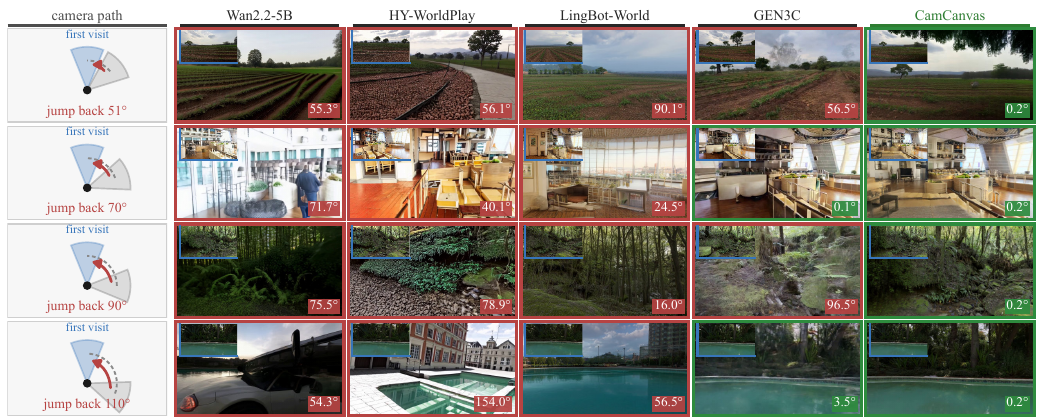}
\caption{\textbf{Panorama jump returns at each jump level.} Each row is one out-of-window return at the jump level shown in the camera-path sketch. For each level, the scene is the one whose CamCanvas return error is closest to the median of that level.}
\label{fig:supp-pano}
\end{figure}

\begin{figure}[!htb]
\centering
\includegraphics[width=\textwidth]{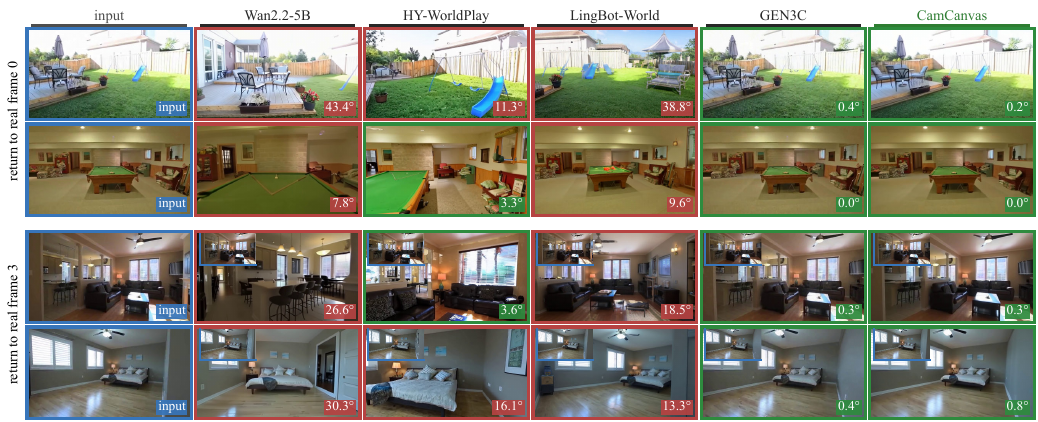}
\caption{\textbf{RealEstate10K jump returns.} The top two clips return to real frame 0, the input, and the bottom two to real frame 3. For each target, the clips are among the five with the highest CamCanvas return PSNR against the real frame.}
\label{fig:supp-re10k}
\end{figure}

\begin{figure}[!htb]
\centering
\includegraphics[width=\textwidth]{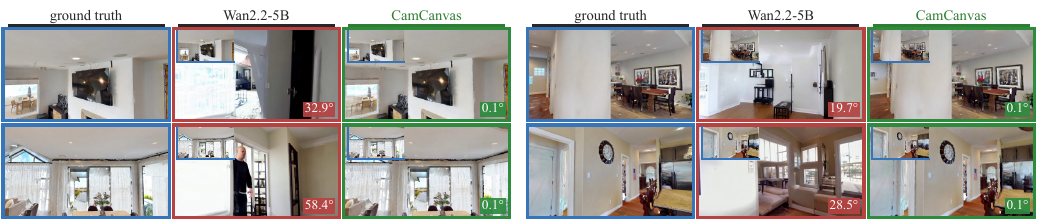}
\caption{\textbf{HM3D returns.} The ground truth is the render at the return pose. The four trajectories come from four buildings and are among the seven with the highest CamCanvas return PSNR against the ground truth.}
\label{fig:supp-hm3d}
\end{figure}

\begin{figure}[!htb]
\centering
\includegraphics[width=\textwidth]{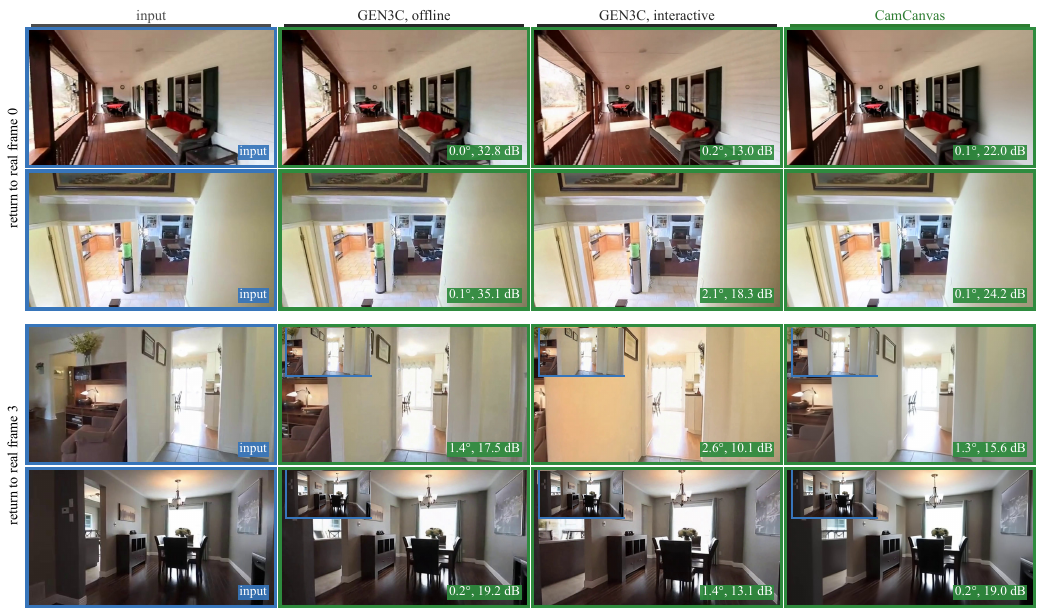}
\caption{\textbf{GEN3C with interactive control.} Offline GEN3C receives the whole trajectory in one pass. Interactive GEN3C receives the outbound poses first and the return poses in a second pass (Table~\ref{tab:gen3c-interactive}). Labels give the return error and the return PSNR against the real frame. For each target, the clips are among the four with the largest drop in return PSNR from offline to interactive GEN3C.}
\label{fig:supp-inter}
\end{figure}

\begin{figure}[!htb]
\centering
\includegraphics[width=\textwidth]{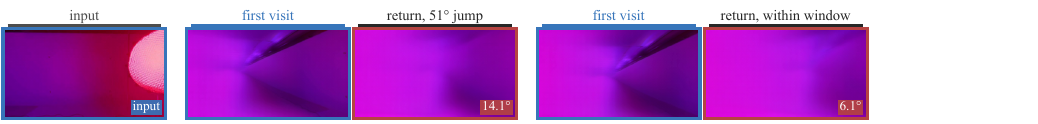}
\caption{\textbf{CamCanvas returns with error above $5^\circ$.} These are the only two of the 182 panorama returns above $5^\circ$, and both come from one scene, a studio lit by a single colored light with little texture. The generated views drift toward a uniform color, and pose estimation on such frames is unreliable.}
\label{fig:supp-fail}
\end{figure}

\paragraph{Qualitative results.}
Figures~\ref{fig:supp-pano}--\ref{fig:supp-fail} add returns that do not appear in Fig.~\ref{fig:qualitative} and follow its conventions: each return shows the method's own first visit as a blue inset and is framed green when its return error is below $5^\circ$ and red otherwise. Each caption states how the examples were chosen. Fig.~\ref{fig:supp-pano} shows one panorama return at each jump level, Fig.~\ref{fig:supp-re10k} RealEstate10K returns to real frames 0 and 3, and Fig.~\ref{fig:supp-hm3d} HM3D returns against ground-truth renders. Fig.~\ref{fig:supp-inter} compares offline and interactive GEN3C with CamCanvas on the clips where interactive control lowers the return PSNR of GEN3C the most.

\paragraph{Failure cases.}
Fig.~\ref{fig:supp-fail} shows every panorama return of CamCanvas with error above $5^\circ$.

\FloatBarrier
\section{Theory and Proofs}
\label{app:theory}

\subsection{Assumptions and Scope}
\label{app:theory:assumptions}
The results below use Definitions~1 and~2 of Sec.~\ref{sec:return:definitions} and the following modeling assumptions. Fix a static pinhole scene, intrinsics $K$, and a command sequence $W_k\in\mathrm{SE}(3)$. For a returning chunk $c$, $y_c^*$ denotes the ideal view at the commanded pose of the content that the model's own first visit represents (A1). The generator has conditional law $p_\theta(z_c\mid z_{\mathcal W_c},u_c)$ of \eqref{eq:chunk-conditional}, with fresh sampling randomness independent of earlier frames given this conditioning set. Write $f$ for the backbone generator of decoded chunks, $p_f$ for the induced conditional law of decoded chunk $x_c$, and $y_c$ for its return frame. Commands and other fixed inputs are suppressed in information-theoretic expressions. If commands depend on generated content, they are included in the conditioning set. Theorem 1 additionally requires $j\notin\mathcal W_c$.

\textbf{A0 (estimator).} Let $d$ be the geodesic metric on $\mathrm{SO}(3)$, so $\angle R=d(\mathrm{Id},R)$. On ideal image pairs, the relative-rotation estimator $E$ has error at most $\epsilon_E$. On the image domain under consideration, it satisfies $d(E(y_1,y),E(y_1,y'))\leq L_E\|y-y'\|$, where the norm is mean absolute RGB difference with intensities in $[0,1]$. The estimator floor is measured with ground-truth renders with the same estimator.

\textbf{A1 (rendering).} On static, jointly visible content with fixed geometry and visibility, the rendered condition $r_c=\Pi(\mathcal S_c,W^{\mathrm{pix}}_{p_c})$ satisfies $\|r_c-y_c^*\|_{\Omega_c}\leq\delta$, where $p_c$ is the pixel frame of $y_c$, $\Omega_c$ is the covered target region and its area fraction is $\kappa\in[0,1]$. The restricted norm averages over covered pixels. Rotation induces $H=KRK^{-1}$, with error from writing and resampling. Translation additionally includes depth error \citep{hartleyzisserman}. For a return scored against the model's own first visit, the ideal target is the view represented by the first observation, so $\delta$ bounds reconstruction relative to that observation.

\textbf{A2 (reproduction and completion).} On covered pixels, $\|y_c-r_c\|_{\Omega_c}\leq\epsilon_I$. On uncovered pixels, the completion differs from the reference view by a mean absolute error of at most $\Delta$. We measure $\epsilon_I$ on the trained adapter (Sec.~\ref{app:theory:constants}). A0 is applied with the first-visit view of A1 as the reference image. The bound of Theorem~2 concerns returns to the same pose. Definition~2 counts a command within its tolerance as a return, but a nearby pose is not the same pose: when the commanded pose differs from the first-visit pose, the angle between them adds to the bound on the return error.

\subsection{Return Blindness}
\label{app:theory:blindness}
\begin{stmt}{Theorem 1}{return blindness}If $f$ is conditioned only on $(z_{\mathcal W_c},u_c)$ and $j\notin\mathcal W_c$, then
\[
 x_c\perp y_j\mid(z_{\mathcal W_c},u_c),\qquad
 I(x_c;y_j)\leq I(z_{\mathcal W_c};y_j).
\]
\end{stmt}
The information inequality is for fixed commands. Return content can reproduce only information about the first visit retained in the window.

\textit{Proof.} The decoded chunk $x_c$ is a deterministic function of $z_c$, so its law given $(z_{\mathcal W_c},u_c)$ follows from \eqref{eq:chunk-conditional}. The conditional law defining $f$ gives $p_f(x_c\mid z_{<c},u_c)=p_f(x_c\mid z_{\mathcal W_c},u_c)$, where $z_{<c}$ denotes the full prior latent history. Including $y_j$ cannot change this law once the window and commands are fixed, proving conditional independence. The resulting Markov chain $y_j\to z_{\mathcal W_c}\to x_c$ yields the information inequality by data processing \citep{coverthomas}. Equivalently, draw $\widetilde y_j$ independently from $p(y_j\mid z_{\mathcal W_c})$ conditional on the window. Replacing $y_j$ by $\widetilde y_j$ leaves the joint distribution with $x_c$ unchanged, and therefore leaves every measurable return-consistency statistic unchanged. Finally, each camera factor in \eqref{eq:prope} relates the target to an available frame. No such factor supplies the missing first-visit appearance as a target view, so the camera input alone cannot supply the first-visit view.

\textit{Scope.} Retained information can still support a correct return: the theorem rules out access to first-visit information beyond that retained information. Attention sinks~\citep{xiao2024streamingllm} or global memory tokens that preserve the first frame violate $j\notin\mathcal W_c$ for $j=0$, where the conditioning set must include all retained memory. Retrieval-based memory restores a first-visit frame to that set, but the retrieved frame was observed at another pose, so the model must still learn the mapping from relative motion to the view \citep{worldmem,contextmemory,worldplay}. Rendered conditions, including geometric rendering in GEN3C \citep{gen3c}, instead supply the view at the commanded pose.

\subsection{Return-Error Bound}
\label{app:theory:bound}
\begin{stmt}{Theorem 2}{return-error bound}Under A0, A1, and A2, with the first-visit frame as the reference (A1),
\[
 e\leq\epsilon_E+L_E(\delta+\epsilon_I)+L_E(1-\kappa)\Delta.
\]
\end{stmt}
The bound has no explicit dependence on jump angle, the number of chunks since the first visit, or window length: same-pose condition tokens make their PRoPE transform the identity, so no term depends on the jump.

\textit{Proof.} Let $y_c^*$ be the ideal same-pose reconstruction of the first-visit reference. A1 and A2 give $\|y_c-y_c^*\|_{\Omega_c}\leq\delta+\epsilon_I$ by the triangle inequality. Splitting the full-image mean over covered and uncovered pixels gives
\[
 \|y_c-y_c^*\|\leq\kappa(\delta+\epsilon_I)+(1-\kappa)\Delta
 \leq\delta+\epsilon_I+(1-\kappa)\Delta.
\]
By A0, $\angle E(y_j,y_c^*)\leq\epsilon_E$, and perturbing the second image changes the estimated rotation by at most $L_E\|y_c-y_c^*\|$. A second triangle inequality gives the bound on $e$.

\subsection{Writing the Canvas}
\label{app:theory:lemmas}
\begin{stmt}{Lemma 1}{written content is fixed}Equation~\ref{eq:write} preserves every written pixel through all subsequent writes. On already covered regions with fixed geometry, visibility, and sampling, repeated reads have a single write/read error $\delta$. Overwriting on every visit admits an additive error recurrence and a linearly accumulating bound.\end{stmt}

\textit{Proof.} If canvas location $s$ is first written at chunk $c_0$, then $M_{c_0+1}(s)=1$. Every subsequent update satisfies $U_c(s)=B_c(s)(1-M_c(s))=0$, so $C_{c+1}(s)=C_c(s)$. Induction preserves the first written value. A fixed read of those locations therefore returns the same samples regardless of intervening chunks. Locations not yet observed may still be written later. With a last-write update, $U_c=B_c$ and the stored value depends on newly generated content. Under a nonexpansive read/write map with additional error at most $\delta'$ per overwrite, the accumulated error satisfies $\delta_c\leq\delta_{c-1}+\delta'$, hence $\delta_n\leq\delta_0+n\delta'$ after $n$ overwrites. With first writes, Theorem~2 applies at every return with the same render error $\delta$, which does not grow with the number of returns. The reproduction error $\epsilon_I$ is measured at each return and may vary.

\subsection{Design of the Condition}
\label{app:theory:design}

\begin{stmt}{Proposition 1}{noise-free condition}Under a Gaussian render-noise model, a condition at timestep zero attains a lower minimum mean squared error of reproduction than a condition noised to the target's timestep.\end{stmt}
\textit{Proof.} On the covered region, the Gaussian render-noise model states that the target latent equals the render up to noise: $z=r+\tau\xi$, $\xi\sim\mathcal N(0,I)$, with $\tau\ge 0$ the noise scale.
The forward process is $z_t=\alpha_t z+\sigma_t\varepsilon$.

\textit{Noise-free condition.} Conditioning on $(z_t,r)$,
\[\mathbb E[z\mid z_t,r]=r+\frac{\alpha_t\tau^2}{\alpha_t^2\tau^2+\sigma_t^2}\,(z_t-\alpha_t r),\qquad
\operatorname{mmse}=\frac{\tau^2\sigma_t^2}{\alpha_t^2\tau^2+\sigma_t^2}\le\tau^2 .\]
As $\tau\to0$ the Bayes-optimal denoiser output is the render $r$ itself.

\textit{Condition supplied at the target's noise level.} Let $r'_t=\alpha_t r+\sigma_t\varepsilon'$, where $\varepsilon'\sim\mathcal N(0,I)$ is independent of $(r,z,z_t)$. This is a degraded version of $r$: $(z,z_t)\to r\to r'_t$ is a Markov chain. Thus
\[
 \mathbb E\|z-\mathbb E[z\mid z_t,r'_t]\|^2
 \ge \mathbb E\|z-\mathbb E[z\mid z_t,r]\|^2.
\]
Conditioning on a degraded observation of $r$ cannot reduce the minimum mean squared error below conditioning on $r$, with equality when $\sigma_t=0$. Under the Gaussian prior $r\sim\mathcal N(\mu,s^2I)$ with $s^2>0$, the inequality is strict for $\sigma_t>0$: $r$ retains positive conditional variance given $(z_t,r'_t)$, and the clean posterior mean depends on $r$ with nonzero coefficient $\sigma_t^2/(\alpha_t^2\tau^2+\sigma_t^2)$.

\textit{Conclusion.} The noise-free condition attains the smallest minimum mean squared error of the two. This reproduction error corresponds to $\epsilon_I$ in A2.

\textit{Scope.} This concerns the Bayes-optimal denoiser under a Gaussian render-noise model, not the trained network.

\begin{stmt}{Proposition 2}{pose alignment}When condition and target share a pose, the target tokens can be formed from the condition tokens by attention weights alone. Otherwise, forming them requires a scene-dependent reprojection.\end{stmt}
\textit{Proof.} With PRoPE, $o_i=\sum_j a_{ij}(P_iP_j^{-1})v_j$.

\textit{Aligned case.} A condition token rendered at the target's commanded pose carries that pose, so $P_iP_j^{-1}=\mathrm{Id}$ and the
condition grid is the target grid. Choosing $a_{ij}$ concentrated on the aligned token gives $o_i=v_{j(i)}$: the target tokens are formed from the condition tokens
by attention weights alone, with no weight encoding the camera motion.

\textit{Misaligned case.} A stored view at pose $W'\neq W$ enters with the fixed transform $P_iP_j^{-1}$, which acts on the pose-carrying
channels. Placing its content into the target view requires the reprojection $\Pi_W\circ\Pi_{W'}^{-1}$, a resampling of token
positions determined by per-pixel depth and the camera motion, so no assignment of attention weights that is independent of the
scene realizes it. The mapping must then be learned in the weights, as for relative commands (Theorem~1).

\subsection{Measured Constants}
\label{app:theory:constants}
\begin{figure}[!ht]
\centering
\includegraphics[width=\textwidth]{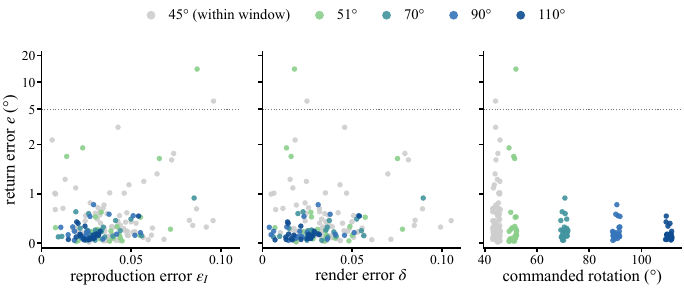}
\caption{\textbf{Return error versus the terms of Theorem~2.} Each point is one of the 182 returns, colored by jump level, plotted against the reproduction error $\epsilon_I$, the render error $\delta$, and the commanded rotation. Target coverage is one throughout.}
\label{fig:bound}
\end{figure}

Figure~\ref{fig:bound} relates the return error to the terms of Theorem~2, and Table~\ref{tab:return-constants} reports the constants.

The constants are measured on 182 returns across 91 panoramas. The canvas is reconstructed from each model's decoded frames preceding the return by keeping the first observation of each location, then rendered at the commanded return poses. Coverage is the rendered validity-mask mean. Reproduction error is covered-region mean absolute difference between output and render. The render error $\delta$ compares the render with the model's own first-visit frame at the target pose, which is the constant of A1. $\delta_{\mathrm{gt}}$ compares the same render with the ground-truth panorama render and also contains the first visit's drift from the commanded sweep. VGGT supplies return error and the ground-truth estimator floor through the same pose pipeline \citep{vggt}.

\begin{table}[H]
\centering
\caption{\textbf{Measured terms of Theorem~2 by jump level.} $n$ counts returns. Image errors are mean absolute differences of $[0,1]$ intensities; $\delta$ is the render error against the model's own frames (A1), and $\delta_{\mathrm{gt}}$ the error against ground-truth renders.}
\label{tab:return-constants}
\small
\begin{tabular}{lrrrrrr}
\toprule
Level & $n$ & $\kappa$ & $\delta$ & $\delta_{\mathrm{gt}}$ & $\epsilon_I$ (median)
& $\epsilon_E$ ($^\circ$) \\
\midrule
\shortstack[l]{Within-window\\return $45^\circ$} & 91 & 1.000 & 0.037 & 0.151 & 0.037 & 0.30 \\
$51^\circ$ & 24 & 1.000 & 0.034 & 0.120 & 0.036 & 0.24 \\
$70^\circ$ & 23 & 1.000 & 0.034 & 0.091 & 0.030 & 0.18 \\
$90^\circ$ & 22 & 1.000 & 0.025 & 0.058 & 0.025 & 0.15 \\
$110^\circ$ & 22 & 1.000 & 0.023 & 0.029 & 0.027 & 0.10 \\
\bottomrule
\end{tabular}
\end{table}

Full coverage removes the completion term for every return in Table~\ref{tab:return-constants}. The return error stays small at every level: its median is $0.16$--$0.31^\circ$, and only 4 of the 182 returns exceed $2^\circ$. Correlations of return error with reproduction error, render error against the model's own frames, ground-truth render discrepancy, and jump angle are $0.32$, $-0.01$, $0.21$, and $-0.14$, respectively.

The two returns above $5^\circ$ have reproduction errors of $0.087$ and $0.096$, among the five largest of all 182 returns, against a median of $0.032$ for returns below $1^\circ$. These errors come from failures to reproduce the supplied condition rather than from the render.

\end{document}